\documentclass{article}

\PassOptionsToPackage{numbers, compress}{natbib}

\usepackage[sort&compress]{natbib}
\usepackage[colorlinks = true, citecolor = blue, linkcolor=blue]{hyperref}
\usepackage[preprint, nonatbib]{neurips_2024}

\usepackage[utf8]{inputenc} 
\usepackage[T1]{fontenc}    
\usepackage{hyperref}       
\usepackage{url}            
\usepackage{booktabs}       
\usepackage{amsfonts}       
\usepackage{nicefrac}       
\usepackage{microtype}      
\usepackage{xcolor}         
\usepackage{wrapfig}

\usepackage{graphicx}
\usepackage{booktabs}
\usepackage{algorithm}
\usepackage{algpseudocode}
\usepackage{multirow}
\usepackage{array}
\usepackage{geometry}
\usepackage{multirow} 
\newcolumntype{P}[1]{>{\centering\arraybackslash}p{#1}}
\usepackage[accsupp]{axessibility}  

\usepackage{orcidlink}

\title{\textsf{Dr-LLaVA}: Visual Instruction Tuning with\\ Symbolic Clinical Grounding}

\author{%
  Shenghuan Sun$^*$\\
  UCSF\\
  \And
  Alexander Schubert$^*$\\
  UC Berkeley \& UCSF\\
  \And
  Gregory M. Goldgof$^*$\\
  MSK Cancer Center\\
  \And
  Zhiqing Sun\\
  CMU\\
  \And
  Thomas Hartvigsen\\
  University of Virginia\\
  \And
  Atul J. Butte\\
  UCSF\\
  \And
  Ahmed Alaa\\
  UC Berkeley \& UCSF
}

\begin{document}

\maketitle

\vspace{-.15in}
\begin{abstract}
Vision-Language Models (VLM) can support clinicians by analyzing medical images and engaging in natural language interactions to~assist~in~diagnostic and treatment tasks. However, VLMs often exhibit ''hallucinogenic'' behavior, generating textual outputs not grounded in contextual multimodal information. This challenge is particularly pronounced in the medical domain, where we do not only require VLM outputs to be accurate in single interactions but also to be consistent with clinical reasoning and diagnostic pathways throughout multi-turn conversations. For this purpose, we propose a new alignment algorithm that uses {\it symbolic representations} of clinical reasoning to ground VLMs in medical knowledge. These representations are utilized to {\bf (i)} generate GPT-4-guided visual instruction tuning data at scale, simulating clinician-VLM conversations with demonstrations of clinical reasoning, and {\bf (ii)} create an automatic reward function that evaluates the clinical validity of VLM generations throughout clinician-VLM interactions. Our algorithm eliminates the need for human involvement in training data generation or reward model construction, reducing costs compared to standard reinforcement learning with human feedback (RLHF). We apply our alignment algorithm to develop \textbf{\texttt{Dr-LLaVA}}, a conversational VLM finetuned for analyzing bone marrow pathology slides, demonstrating strong performance in multi-turn medical conversations.

{\small {\bf Code:} \href{https://github.com/AlaaLab/Dr-LLaVA}{Link}}; {\small {\bf Demo:} \href{https://huggingface.co/spaces/alaa-lab/Dr-LLaVA}{Link}}
\end{abstract}

\vspace{-.15in}
\section{Introduction}
\label{sec:intro}


Vision-language models (VLMs) \cite{radford2021,liu2024visual,zhang2023llavar}, which integrate large language models (LLMs) \cite{brown2020language,touvron2023llama, touvron2023llama-b,openai_chatgpt,openai_gpt4_2023} with vision encoders, have demonstrated strong capabilities in answering complex questions that require both visual and textual reasoning. In the medical domain, VLMs hold great promise-they could~serve~as~helpful assistants for clinicians, researchers, and trainees, providing an interactive natural language interface for the analysis of medical images within clinical workflows \cite{li2023llava, monajatipoor2022berthop, chen2023generative, lu2023towards, naseem2022vision, lu2023foundational}. However, the practical utility of present VLMs is significantly limited by their tendency to ``hallucinate''. In this context, hallucination refers not only to instances where the model generates responses ungrounded in visual input but also to cases where, in multi-turn interactions, its responses are incoherent, contradictory, or misaligned with diagnostic pathways and domain knowledge.


The currently predominant methods to reduce hallucinations in VLMs such as Reinforcement Learning from Human Feedback (RLHF) \cite{yu2023rlhf, stiennon2020learning, ouyang2022training, sun2023aligning} are not well-suited for the multimodal medical context. Using RLHF to align VLMs with visually-grounded clinical reasoning requires multimodal training data showcasing the reasoning process within multi-turn QA dialogues. These datasets are not readily available in health systems. Synthesizing these datasets and collecting clinician feedback on VLM responses is bottle-necked by the expertise of medical professionals. Unlike the LLaVA-RLHF model in \cite{sun2023aligning}, which gathered human feedback from non-expert crowdworkers for simple, common-sense visual QA tasks, this process cannot be scaled without the involvement of clinicians. Due to these limitations, specialized medical VLMs like LLaVA-Med \cite{li2023llava} and PathChat \cite{liu2024visual} have been confined to supervised finetuning, relying on automatically generated QA tasks using image captions. Moreover, both existing general-purpose and medical VLMs have only been finetuned for single-turn QA, rather than for multi-turn conversations that convey complex and interactive clinical reasoning.


\vspace{-.05in}
\begin{figure}[H]
\begin{minipage}{0.5\linewidth} 
In this work, we capitalize on the key insight that many clinical reasoning processes can be formalized as a hierarchical set of {\it symbolic rules}. This enables the decomposition of ambiguous medical inquiries into a sequence of logical steps, where the outcomes of earlier sub-analyses constrain the set of permissible diagnoses in subsequent stages. Our proposed method leverages these rules~to automatically synthesize a realistic multi-turn VLM-clinician conversation finetuning dataset.  Furthermore, we design a novel alignment algorithm that extends the RLHF procedure by introducing a reward function that automatically evaluates VLM responses, promoting accurate single-turn responses while ensuring consistency with correct clinical reasoning across the entire multi-turn dialogue. 
\end{minipage}
\enspace 
\begin{minipage}{0.50\linewidth}
\centering
\includegraphics[width=2.7in]{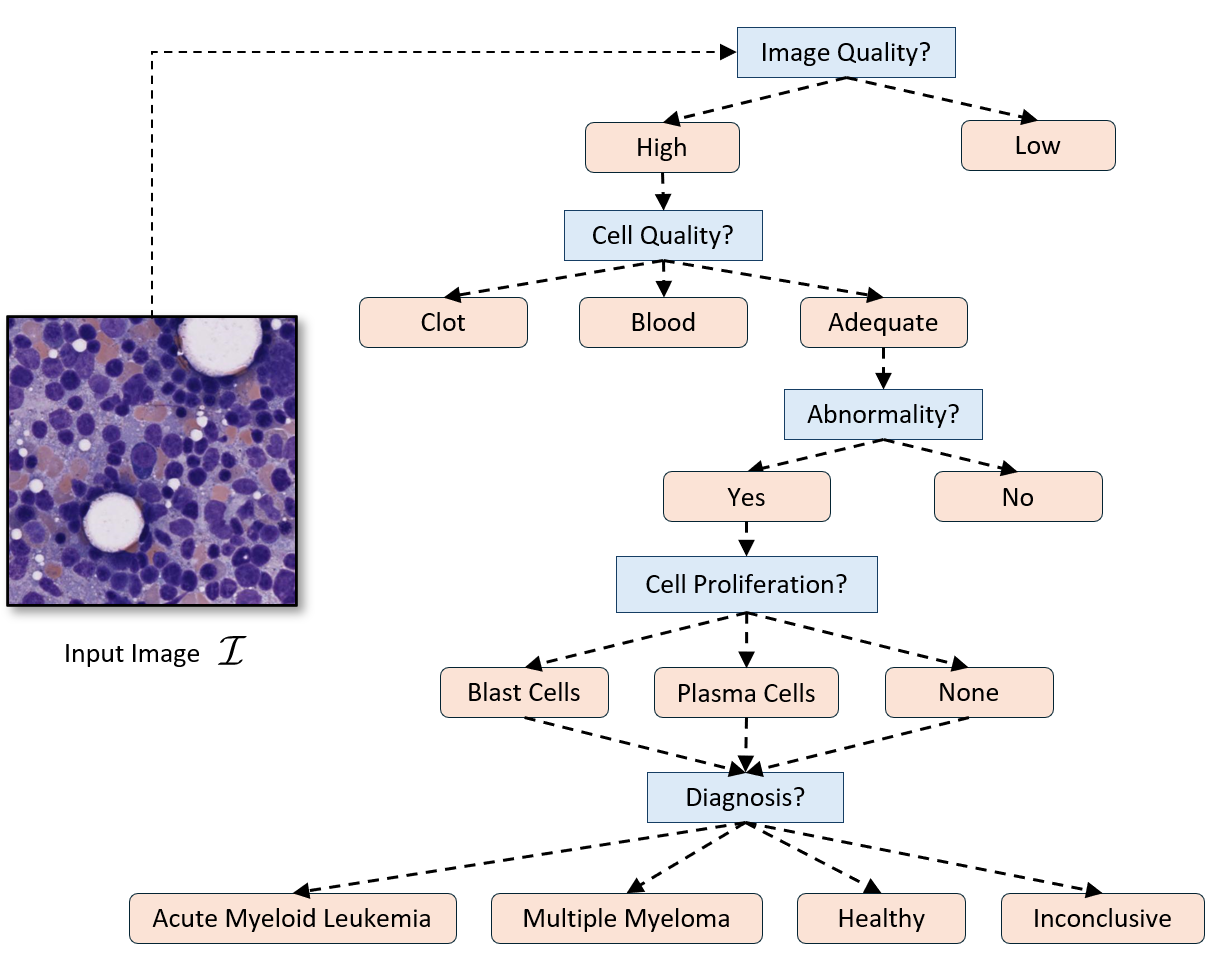}
\vspace{-.1in}
\caption{{\footnotesize Symbolic representation of clinical reasoning in blood cancer diagnosis.}}
\label{fig:logic}
\end{minipage}
\end{figure}
\vspace{-.26in}

 This enables us to adapt VLMs to multi-turn imaging-based conversational diagnostic tasks, while eliminating the need~for~human~involvement in training data generation or feedback collection.

We demonstrate the utility of our proposed algorithm by finetuning the LLaVA model \cite{liu2024visual} to develop \textbf{\texttt{Dr-LLaVA}}, a VLM designed for diagnosing blood cancer using bone marrow pathology images. To this end, we curated a dataset comprising 16,340 bone marrow image patches and generate corresponding multi-turn clinician-VLM conversations. Our results show that \textbf{\texttt{Dr-LLaVA}} outperforms state-of-the-art VLMs in both single- and multi-turn conversational settings. Furthermore, ablation experiments show that our instruction-tuning framework enabled \textbf{\texttt{Dr-LLaVA}} to attain high robustness to variations in question sequencing and to outperform other baselines in identifying and correcting erroneous information in clinician prompts. These findings underscore the value of integrating clinical domain knowledge into fine-tuning approaches using a hybrid symbolic and data-driven method, thereby developing trustworthy and accurate conversational assistants in medicine.

\vspace{-.15in}
\section{Visual Instruction Tuning with Symbolic Clinical Grounding}

\begin{figure*}[t]
\centering
  \includegraphics[width=\textwidth]{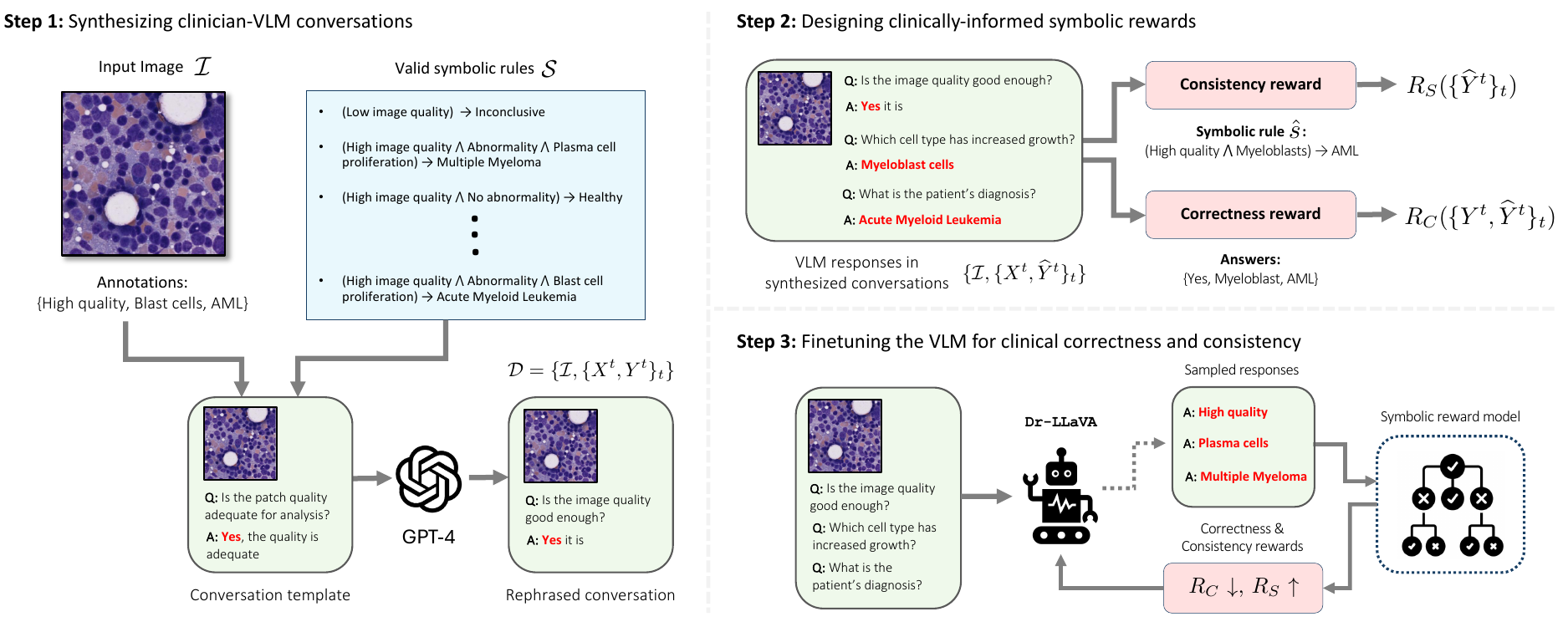}
  \vspace{-.15in}
  \caption{
  {\footnotesize \textbf{Pictorial depiction of the \textbf{\texttt{Dr-LLaVA}} training pipeline} (a) Multi-turn conversations consistent with symbolic clinical reasoning are generated for each medical image, utilizing GPT-4 for diverse phrasing. (b) A symbolic reward function evaluates VLM responses, checking individual correctness and clinical validity. (c) Using the dataset from (a) and the reward model from (b), a pretrained VLM is finetuned via RL.}}
  
  \label{fig:Dr-LLaVA-framework}
  \vspace{.025in}
  \hrule
  \vspace{-.15in}
\end{figure*}

Many medical diagnostic processes can be described using a relatively small number of logical rules applied sequentially. Fig. \ref{fig:logic} presents such a symbolic representation, constructed and adjudicated by an expert pathologist, which outlines each step in the process for diagnosing blood cancer based on bone marrow pathology slides. The diagnostic process encompasses key steps such as evaluating image quality, verifying the presence of sufficient nucleated cells, and detecting abnormalities to establish a diagnosis. The decision tree delineates the valid reasoning pathways that VLM responses must adhere to in order to maintain clinical coherence. For example, it would be invalid if a slide deemed too low in quality for assessment was still used for diagnosis. Formally, we define a set of symbolic rules, $\mathcal{S}$, which outlines all valid reasoning paths in the decision tree. Our instruction tuning framework leverages this symbolic representation to (a) synthesize a dataset of clinician-VLM conversations, (b) automatically evaluate the clinical consistency of VLM responses, and (c) finetune the VLM to ensure clinical correctness and coherence. (See Fig. \ref{fig:Dr-LLaVA-framework} for a pictorial depiction.)


\textbf{Step 1: Synthesizing clinician-VLM conversations.} We synthesize clinician-VLM conversations using a dataset derived from bone~marrow~aspirate~(BMA) whole slide images, annotated by hematopathologists and sourced from the clinical archives of an academic medical center. The dataset includes images indicative of various conditions: blood contamination, particle-enriched contamination, acute myeloid leukemia, multiple myeloma, and healthy states. For each image, we use the hematopathologist's annotations to select the symbolic rule from $\mathcal{S}$ that describe the corresponding diagnostic analysis. These rules are then used to construct a multimodal instruction tuning dataset $\mathcal{D} = {(\mathcal{I}_i, {X^t_i, Y^t_i}_t)}_i$, where each image $\mathcal{I}_i$ is paired with multi-turn clinical conversations ${X^t_i, Y^t_i}_t$. Each $X_i^t$ represents the $t$-th clinician prompt, and $Y_i^t$ is the corresponding target response, generated by applying textual templates to the image annotations for the respective analysis step (e.g., image quality, cell types, diagnoses). The conversations consist of five question-answer pairs that follow the diagnostic steps of the symbolic rule. To introduce diversity, GPT-4 is used to generate paraphrased prompts and responses. An illustration of this dataset synthesis process is provided in Fig. \ref{fig:Dr-LLaVA-framework}. This dataset serves as the basis for our instruction-tuning framework, which combines supervised finetuning and reinforcement learning. (More details are provided in the appendix.)

\textbf{Step 2: Designing clinically-informed symbolic rewards.} In contrast to standard RLHF approaches that rely on human feedback to evaluate ambiguous qualities of model outputs \cite{sun2023aligning,ouyang2022training,sun2024aligning}, our conversational diagnostic system leverages symbolic representations (Fig. \ref{fig:logic}) to convert complex diagnostic questions into a sequence of discrete decisions. This approach enables us to define an efficient keyword-matching algorithm that evaluates VLM responses against specific terms associated with a limited set of admissible answer categories in the decision tree. This facilitates automated evaluation without costly human annotation. A comprehensive list of keywords is provided in the appendix.

Given this discrete categorization, we define a reward model that assesses both the correctness of model responses and their alignment with valid clinical reasoning. For an input image $\mathcal{I}_i$ and a sequence of prompted VLM outputs ${(X^t_i, \widehat{Y}^t_i)}_t$, we compute the reward function as:
\vspace{-0.1in}
\begin{equation*}
R((\widehat{Y}^t_i, Y^t_i),\ldots, (\widehat{Y}^T_i, Y^T_i)) = \frac{1}{T}\sum_{t=1}^{T} \underbrace{R_C(Y^t_i, \widehat{Y}^t_i)}_{\mbox{\scriptsize {\bf Correctness} of responses}} + \underbrace{\lambda \cdot R_{\mathcal{S}}(\{\widehat{Y}^t_i\}_t)}_{\mbox{\scriptsize {\bf Consistency} with valid reasoning}} + R_l - R_m
\label{reward_eq}
\end{equation*}
Here, $R_C$ evaluates the accuracy of individual model responses against ground truth, while $R_{\mathcal{S}}(.)$ assesses whether the VLM's answer sequence aligns with a clinically valid reasoning path. The hyperparameter $\lambda$ balances correctness and consistency rewards. In addition, following \cite{sun2023aligning}, we include further terms $R_m$ to penalize ambiguous responses and $R_l$ to discourage significant deviations between the length of the VLM's answer and the target answer length.

\textbf{Step 3: Finetuning the VLM for clinical correctness and consistency.} We employ a two-stage approach to optimize the VLM for clinical tasks. First, we perform supervised finetuning (SFT) to obtain $\pi_{\text{SFT}}^{\phi}$. Subsequently, we further refine this model using Reinforcement Learning (RL) based on our automatically evaluated symbolic rewards.
In the RL stage, we treat $\pi_{\text{SFT}}^{\phi}$ as our initial policy model, training it to generate accurate responses to clinical queries that maximize the reward model output. Following \cite{ouyang2022training, sun2023aligning}, we implement Proximal Policy Optimization (PPO) \cite{schulman2017proximal} with a per-token Kullback-Leibler (KL) penalty to mitigate reward hacking. This penalty constrains the RL-tuned model's divergence from the SFT model.
Given a dataset of medical images, clinical analysis prompts, and their respective answers $\mathcal{D}_{RL} = \{(\mathcal{I}_i, \{X^t_i, Y^t_i\}_t)\}_i$ we define the full finetuning loss as:


\begin{equation}
    \mathcal{L}(\pi_{\text{RL}}^{\phi}) = -\mathbb{E}_{(\mathcal{I},X, Y) \in \mathcal{D}_{\text{RL}}, \widehat{Y} \sim \pi_{\text{RL}}(\hat{Y}|\mathcal{I},X)} \left[ R(\{\widehat{Y}^t, Y^t\}_t)
    - \beta \cdot D_{\text{KL}}(\pi_{\text{RL}}^{\phi}(\widehat{Y}|\mathcal{I},X) \| \pi_{\text{SFT}}^{\phi}(\widehat{Y}|\mathcal{I},X)) \right] \nonumber
\end{equation}

Notably, unlike previous RLHF methods \cite{sun2023aligning}, our loss function $\mathcal{L}(\pi_{\text{RL}}^{\phi})$ is computed based on the entire multi-turn conversation. since the consistency reward in (\ref{reward_eq}), which evaluates the sequence of all model responses collectively.

\vspace{-.15in}
\section{Results}
\label{sec:blind}
\textbf{Baselines and Evaluation Metrics.} We employ our finetuning algorithm to develop \textbf{\texttt{Dr-LLaVA}}, a conversational VLM specialized in bone marrow pathology slide analysis. We evaluate \textbf{\texttt{Dr-LLaVA}} against state-of-the-art VLMs including LLaVA \cite{liu2024visual}, OpenFlamingo \cite{awadalla2023openflamingo}, MiniGPT-4 \cite{zhu2023minigpt}, and LLaMA-Adapter \cite{zhang2023llama}. Given the limited zero-shot performance in this specialized domain, all models undergo supervised finetuning on our synthesized conversational dataset for four epochs prior to evaluation on a 20\% holdout test set. Detailed training specifications are provided in the appendix. We evaluate model performance using three metrics: Question-level Accuracy ($A_Q$), Conversation-level Accuracy ($A_C$), and Diagnostic Accuracy ($A_D$). $A_Q$ measures the proportion of correctly answered questions across all conversations, while $A_C$ represents the \textit{fraction of conversations} where \textit{all questions} were answered correctly. $A_D$ assesses the model's ability to make a correct final diagnosis, independent of its performance in preceding analysis steps. 

\begin{figure}[t]
\begin{minipage}{.5\textwidth}
\centering
\scalebox{0.9}{ 
\footnotesize
\begin{tabular}{@{}lccc@{}}
\toprule
 & \multicolumn{3}{c}{{\bf Metrics}} \\
{\bf Baseline} & \textbf{$A_Q$} & \textbf{$A_C$} & \textbf{$A_D$} \\
\toprule
{LLaVA-0-shot \cite{liu2024visual}} & 16.5 & 0.0 & 12.6  \\
{OpenFlamingo-SFT \cite{awadalla2023openflamingo}} &60.5 &31.3 &55.8 \\
{LLaMA-Adapter-SFT \cite{zhang2023llama}} & 68.6 & 37.8 & 70.2 \\
{MiniGPT-4-SFT \cite{zhu2023minigpt}} & 64.1 & 32.9 & 50.0 \\
{LLaVA-Med-SFT \cite{li2023llava}} & 78.2 & 55.6 & 76.5 \\
{LLaVA-SFT \cite{liu2024visual}} & 77.4 & 47.6 & 77.3  \\
\textbf{\texttt{Dr-LLaVA}} & \textbf{89.6} & \textbf{70.0} & \textbf{84.7} \\
\hline
\end{tabular}}

\vspace{.035in}
\text{\scriptsize Table 1: Performance in single-turn conversations.}

\end{minipage}
\begin{minipage}{.5\textwidth}
\vspace{.1in} 
\centering
\scalebox{0.9}{ 
\footnotesize
\begin{tabular}{@{}lccc@{}}
\toprule
 & \multicolumn{3}{c}{{\bf Metrics}} \\
{\bf Baseline} & \textbf{$A_Q$} & \textbf{$A_C$} & \textbf{$A_D$} \\
\toprule
{LLaMA-Adapter-SFT \cite{zhang2023llama}} & 70.4 & 42.5 & 75.4 \\
{MiniGPT-4-SFT \cite{zhu2023minigpt}} & 75.8 & 44.2 & 75.4 \\
{OpenFlamingo-SFT \cite{awadalla2023openflamingo}} & 81.4 & 46.4 & 69.9 \\
{LLaVA-Med-SFT \cite{li2023llava}} & 91.2 & 85.6 & 90.3 \\
{LLaVA-SFT \cite{liu2024visual}} & 92.4 & 90.1 & 91.8 \\
\textbf{\texttt{Dr-LLaVA}} & \textbf{93.6} & \textbf{90.8} & \textbf{92.0} \\
\hline
\end{tabular}}

\vspace{.035in}
\text{\scriptsize Table 2: Performance in multi-turn conversations.}

\end{minipage}
\vspace{-0.15in}
\end{figure}

\textbf{Performance.} We first evaluate \textbf{\texttt{Dr-LLaVA}} in single-question scenarios, where a clinician seeks clarification on a specific step in the image analysis process without prior conversational context. Table~1 demonstrates that our finetuning algorithm significantly enhances \textbf{\texttt{Dr-LLaVA}}'s performance across all metrics, surpassing state-of-the-art VLMs. Notably, \textbf{\texttt{Dr-LLaVA}} achieves a Question-level Accuracy of 89.6\%, substantially higher than the top baseline model, LLaVA-SFT. Moreover, \textbf{\texttt{Dr-LLaVA}} exhibits a 13 percentage point increase in Conversation-level Accuracy over the best baseline, even without conversational context, highlighting the efficacy of our finetuning algorithm in ensuring clinically consistent answers. The challenges of zero-shot generalization to this specialized domain are evident, with the general LLaVA model's performance falling below 20\% across all metrics. We further evaluate \textbf{\texttt{Dr-LLaVA}} in a conversational context, with results presented in Table~2. While access to conversational context generally improve performance across compared models, \textbf{\texttt{Dr-LLaVA}} consistently outperforms across all three metrics. This superior performance underscores \textbf{\texttt{Dr-LLaVA}}'s advanced adaptive reasoning capabilities, demonstrating its proficiency in extracting and utilizing critical information from conversational contexts. Further results in the appendix demonstrate \textbf{\texttt{Dr-LLaVA}}'s robustness to varied conversation sequences and misleading clinician prompts, and the critical impact of each reward component on model performance and behavior.

\vspace{-.15in}
\section{Conclusion}
Vision-language models (VLMs) hold the potential of becoming valuable tools for clinicians, researchers, and trainees, offering an interactive natural language interface for medical image analysis within clinical workflows. Yet, their utility is often compromised by the generation of “hallucinated” outputs that deviate from sound medical reasoning, leading to a lack of trust in their responses. This paper presents a novel alignment algorithm designed to finetune VLMs, grounding them in the symbolic representations of medical image analysis processes. This approach ensures the production of clinically valid and consistent responses throughout multi-turn interactions. Applying this algorithm, we developed Dr-LLaVA, a VLM specifically tailored for analyzing bone marrow image patches. Our findings indicate that Dr-LLaVA not only performs well in straightforward question-answer scenarios but also exhibits superior adaptability and accuracy in intricate, multi-turn clinical dialogues, surpassing other advanced VLMs. These outcomes highlight the critical role of precise model alignment with medical knowledge, in order to make VLMs more reliable and effective in supporting decision-making processes in specialist domains.

\newpage

\par\vfill\par
\clearpage  

%
%
\bibliographystyle{unsrt}
\bibliography{egbib}

\begin{thebibliography}{10}

\bibitem{radford2021}
Alec Radford, Jong~Wook Kim, Chris Hallacy, Aditya Ramesh, Gabriel Goh, Sandhini Agarwal, Girish Sastry, Amanda Askell, Pamela Mishkin, Jack Clark, et~al.
\newblock Learning transferable visual models from natural language supervision.
\newblock In {\em International conference on machine learning}, pages 8748--8763. PMLR, 2021.

\bibitem{liu2024visual}
Haotian Liu, Chunyuan Li, Qingyang Wu, and Yong~Jae Lee.
\newblock Visual instruction tuning.
\newblock {\em Advances in neural information processing systems}, 36, 2024.

\bibitem{zhang2023llavar}
Yanzhe Zhang, Ruiyi Zhang, Jiuxiang Gu, Yufan Zhou, Nedim Lipka, Diyi Yang, and Tong Sun.
\newblock Llavar: Enhanced visual instruction tuning for text-rich image understanding.
\newblock {\em arXiv preprint arXiv:2306.17107}, 2023.

\bibitem{brown2020language}
Tom Brown, Benjamin Mann, Nick Ryder, Melanie Subbiah, Jared~D Kaplan, Prafulla Dhariwal, Arvind Neelakantan, Pranav Shyam, Girish Sastry, Amanda Askell, et~al.
\newblock Language models are few-shot learners.
\newblock {\em Advances in neural information processing systems}, 33:1877--1901, 2020.

\bibitem{touvron2023llama}
Hugo Touvron, Thibaut Lavril, Gautier Izacard, Xavier Martinet, Marie-Anne Lachaux, Timoth{\'e}e Lacroix, Baptiste Rozi{\`e}re, Naman Goyal, Eric Hambro, Faisal Azhar, et~al.
\newblock Llama: Open and efficient foundation language models.
\newblock {\em arXiv preprint arXiv:2302.13971}, 2023.

\bibitem{touvron2023llama-b}
Hugo Touvron, Louis Martin, Kevin Stone, Peter Albert, Amjad Almahairi, Yasmine Babaei, Nikolay Bashlykov, Soumya Batra, Prajjwal Bhargava, Shruti Bhosale, et~al.
\newblock Llama 2: Open foundation and fine-tuned chat models, 2023.
\newblock {\em URL https://arxiv. org/abs/2307.09288}, 2023.

\bibitem{openai_chatgpt}
{OpenAI}.
\newblock Chatgpt: Optimizing language models for dialogue.
\newblock \url{https://openai.com/blog/chatgpt}, 2022.
\newblock Accessed: [Your Access Date].

\bibitem{openai_gpt4_2023}
OpenAI.
\newblock Gpt-4 technical report.
\newblock {\em arXiv}, 2023.
\newblock Accessed: [Your Access Date].

\bibitem{li2023llava}
Chunyuan Li, Cliff Wong, Sheng Zhang, Naoto Usuyama, Haotian Liu, Jianwei Yang, Tristan Naumann, Hoifung Poon, and Jianfeng Gao.
\newblock Llava-med: Training a large language-and-vision assistant for biomedicine in one day.
\newblock {\em arXiv preprint arXiv:2306.00890}, 2023.

\bibitem{monajatipoor2022berthop}
Masoud Monajatipoor, Mozhdeh Rouhsedaghat, Liunian~Harold Li, C-C Jay~Kuo, Aichi Chien, and Kai-Wei Chang.
\newblock Berthop: An effective vision-and-language model for chest x-ray disease diagnosis.
\newblock In {\em International Conference on Medical Image Computing and Computer-Assisted Intervention}, pages 725--734. Springer, 2022.

\bibitem{chen2023generative}
Yinda Chen, Che Liu, Wei Huang, Sibo Cheng, Rossella Arcucci, and Zhiwei Xiong.
\newblock Generative text-guided 3d vision-language pretraining for unified medical image segmentation.
\newblock {\em arXiv preprint arXiv:2306.04811}, 2023.

\bibitem{lu2023towards}
Ming~Y Lu, Bowen Chen, Drew~FK Williamson, Richard~J Chen, Ivy Liang, Tong Ding, Guillaume Jaume, Igor Odintsov, Andrew Zhang, Long~Phi Le, et~al.
\newblock Towards a visual-language foundation model for computational pathology.
\newblock {\em arXiv preprint arXiv:2307.12914}, 2023.

\bibitem{naseem2022vision}
Usman Naseem, Matloob Khushi, and Jinman Kim.
\newblock Vision-language transformer for interpretable pathology visual question answering.
\newblock {\em IEEE Journal of Biomedical and Health Informatics}, 27(4):1681--1690, 2022.

\bibitem{lu2023foundational}
Ming~Y Lu, Bowen Chen, Drew~FK Williamson, Richard~J Chen, Kenji Ikamura, Georg Gerber, Ivy Liang, Long~Phi Le, Tong Ding, Anil~V Parwani, et~al.
\newblock A foundational multimodal vision language ai assistant for human pathology.
\newblock {\em arXiv preprint arXiv:2312.07814}, 2023.

\bibitem{yu2023rlhf}
Tianyu Yu, Yuan Yao, Haoye Zhang, Taiwen He, Yifeng Han, Ganqu Cui, Jinyi Hu, Zhiyuan Liu, Hai-Tao Zheng, Maosong Sun, et~al.
\newblock Rlhf-v: Towards trustworthy mllms via behavior alignment from fine-grained correctional human feedback.
\newblock {\em arXiv preprint arXiv:2312.00849}, 2023.

\bibitem{stiennon2020learning}
Nisan Stiennon, Long Ouyang, Jeffrey Wu, Daniel Ziegler, Ryan Lowe, Chelsea Voss, Alec Radford, Dario Amodei, and Paul~F Christiano.
\newblock Learning to summarize with human feedback.
\newblock {\em Advances in Neural Information Processing Systems}, 33:3008--3021, 2020.

\bibitem{ouyang2022training}
Long Ouyang, Jeffrey Wu, Xu~Jiang, Diogo Almeida, Carroll Wainwright, Pamela Mishkin, Chong Zhang, Sandhini Agarwal, Katarina Slama, Alex Ray, et~al.
\newblock Training language models to follow instructions with human feedback.
\newblock {\em Advances in Neural Information Processing Systems}, 35:27730--27744, 2022.

\bibitem{sun2023aligning}
Zhiqing Sun, Sheng Shen, Shengcao Cao, Haotian Liu, Chunyuan Li, Yikang Shen, Chuang Gan, Liang-Yan Gui, Yu-Xiong Wang, Yiming Yang, et~al.
\newblock Aligning large multimodal models with factually augmented rlhf.
\newblock {\em arXiv preprint arXiv:2309.14525}, 2023.

\bibitem{sun2024aligning}
Shenghuan Sun, Greg Goldgof, Atul Butte, and Ahmed~M Alaa.
\newblock Aligning synthetic medical images with clinical knowledge using human feedback.
\newblock {\em Advances in Neural Information Processing Systems}, 36, 2024.

\bibitem{schulman2017proximal}
John Schulman, Filip Wolski, Prafulla Dhariwal, Alec Radford, and Oleg Klimov.
\newblock Proximal policy optimization algorithms.
\newblock {\em arXiv preprint arXiv:1707.06347}, 2017.

\bibitem{awadalla2023openflamingo}
Anas Awadalla, Irena Gao, Josh Gardner, Jack Hessel, Yusuf Hanafy, Wanrong Zhu, Kalyani Marathe, Yonatan Bitton, Samir Gadre, Shiori Sagawa, et~al.
\newblock Openflamingo: An open-source framework for training large autoregressive vision-language models.
\newblock {\em arXiv preprint arXiv:2308.01390}, 2023.

\bibitem{zhu2023minigpt}
Deyao Zhu, Jun Chen, Xiaoqian Shen, Xiang Li, and Mohamed Elhoseiny.
\newblock Minigpt-4: Enhancing vision-language understanding with advanced large language models.
\newblock {\em arXiv preprint arXiv:2304.10592}, 2023.

\bibitem{zhang2023llama}
Renrui Zhang, Jiaming Han, Aojun Zhou, Xiangfei Hu, Shilin Yan, Pan Lu, Hongsheng Li, Peng Gao, and Yu~Qiao.
\newblock Llama-adapter: Efficient fine-tuning of language models with zero-init attention.
\newblock {\em arXiv preprint arXiv:2303.16199}, 2023.

\bibitem{liu2023improved}
Haotian Liu, Chunyuan Li, Yuheng Li, and Yong~Jae Lee.
\newblock Improved baselines with visual instruction tuning.
\newblock {\em arXiv preprint arXiv:2310.03744}, 2023.

\bibitem{chiang2023vicuna}
Wei-Lin Chiang, Zhuohan Li, Zi~Lin, Ying Sheng, Zhanghao Wu, Hao Zhang, Lianmin Zheng, Siyuan Zhuang, Yonghao Zhuang, Joseph~E Gonzalez, et~al.
\newblock Vicuna: An open-source chatbot impressing gpt-4 with 90\%* chatgpt quality.
\newblock {\em See https://vicuna. lmsys. org (accessed 14 April 2023)}, 2023.

\bibitem{radford2021learning}
Alec Radford, Jong~Wook Kim, Chris Hallacy, Aditya Ramesh, Gabriel Goh, Sandhini Agarwal, Girish Sastry, Amanda Askell, Pamela Mishkin, Jack Clark, et~al.
\newblock Learning transferable visual models from natural language supervision.
\newblock In {\em International conference on machine learning}, pages 8748--8763. PMLR, 2021.

\bibitem{dubois2023alpacafarm}
Yann Dubois, Xuechen Li, Rohan Taori, Tianyi Zhang, Ishaan Gulrajani, Jimmy Ba, Carlos Guestrin, Percy Liang, and Tatsunori~B Hashimoto.
\newblock Alpacafarm: A simulation framework for methods that learn from human feedback.
\newblock {\em arXiv preprint arXiv:2305.14387}, 2023.

\bibitem{chowdhery2023palm}
Aakanksha Chowdhery, Sharan Narang, Jacob Devlin, Maarten Bosma, Gaurav Mishra, Adam Roberts, Paul Barham, Hyung~Won Chung, Charles Sutton, Sebastian Gehrmann, et~al.
\newblock Palm: Scaling language modeling with pathways.
\newblock {\em Journal of Machine Learning Research}, 24(240):1--113, 2023.

\bibitem{anil2023palm}
Rohan Anil, Andrew~M Dai, Orhan Firat, Melvin Johnson, Dmitry Lepikhin, Alexandre Passos, Siamak Shakeri, Emanuel Taropa, Paige Bailey, Zhifeng Chen, et~al.
\newblock Palm 2 technical report.
\newblock {\em arXiv preprint arXiv:2305.10403}, 2023.

\bibitem{workshop2022bloom}
BigScience Workshop, Teven~Le Scao, Angela Fan, Christopher Akiki, Ellie Pavlick, Suzana Ili{\'c}, Daniel Hesslow, Roman Castagn{\'e}, Alexandra~Sasha Luccioni, Fran{\c{c}}ois Yvon, et~al.
\newblock Bloom: A 176b-parameter open-access multilingual language model.
\newblock {\em arXiv preprint arXiv:2211.05100}, 2022.

\bibitem{muennighoff2022crosslingual}
Niklas Muennighoff, Thomas Wang, Lintang Sutawika, Adam Roberts, Stella Biderman, Teven~Le Scao, M~Saiful Bari, Sheng Shen, Zheng-Xin Yong, Hailey Schoelkopf, et~al.
\newblock Crosslingual generalization through multitask finetuning.
\newblock {\em arXiv preprint arXiv:2211.01786}, 2022.

\bibitem{taori2023stanford}
Rohan Taori, Ishaan Gulrajani, Tianyi Zhang, Yann Dubois, Xuechen Li, Carlos Guestrin, Percy Liang, and Tatsunori~B Hashimoto.
\newblock Stanford alpaca: An instruction-following llama model, 2023.

\bibitem{unknown-author-2023}
{Introducing Claude}, 3 2023.

\bibitem{su2019vl}
Weijie Su, Xizhou Zhu, Yue Cao, Bin Li, Lewei Lu, Furu Wei, and Jifeng Dai.
\newblock Vl-bert: Pre-training of generic visual-linguistic representations.
\newblock {\em arXiv preprint arXiv:1908.08530}, 2019.

\bibitem{ramesh2021zero}
Aditya Ramesh, Mikhail Pavlov, Gabriel Goh, Scott Gray, Chelsea Voss, Alec Radford, Mark Chen, and Ilya Sutskever.
\newblock Zero-shot text-to-image generation.
\newblock In {\em International conference on machine learning}, pages 8821--8831. Pmlr, 2021.

\bibitem{alayrac2022flamingo}
Jean-Baptiste Alayrac, Jeff Donahue, Pauline Luc, Antoine Miech, Iain Barr, Yana Hasson, Karel Lenc, Arthur Mensch, Katherine Millican, Malcolm Reynolds, et~al.
\newblock Flamingo: a visual language model for few-shot learning.
\newblock {\em Advances in Neural Information Processing Systems}, 35:23716--23736, 2022.

\bibitem{byra2023few}
Michal Byra, Muhammad~Febrian Rachmadi, and Henrik Skibbe.
\newblock Few-shot medical image classification with simple shape and texture text descriptors using vision-language models.
\newblock {\em arXiv preprint arXiv:2308.04005}, 2023.

\bibitem{liu2023qilin}
Junling Liu, Ziming Wang, Qichen Ye, Dading Chong, Peilin Zhou, and Yining Hua.
\newblock Qilin-med-vl: Towards chinese large vision-language model for general healthcare.
\newblock {\em arXiv preprint arXiv:2310.17956}, 2023.

\bibitem{huang2023visual}
Zhi Huang, Federico Bianchi, Mert Yuksekgonul, Thomas~J Montine, and James Zou.
\newblock A visual--language foundation model for pathology image analysis using medical twitter.
\newblock {\em Nature medicine}, 29(9):2307--2316, 2023.

\bibitem{huang2021gloria}
Shih-Cheng Huang, Liyue Shen, Matthew~P Lungren, and Serena Yeung.
\newblock Gloria: A multimodal global-local representation learning framework for label-efficient medical image recognition.
\newblock In {\em Proceedings of the IEEE/CVF International Conference on Computer Vision}, pages 3942--3951, 2021.

\bibitem{wu2023towards}
Chaoyi Wu, Xiaoman Zhang, Ya~Zhang, Yanfeng Wang, and Weidi Xie.
\newblock Towards generalist foundation model for radiology.
\newblock {\em arXiv preprint arXiv:2308.02463}, 2023.

\bibitem{shu2023visual}
Chang Shu, Fu~Liu, and Collier Shareghi.
\newblock Visual med-alpaca: A parameter-efficient biomedical llm with visual capabilities, 2023.

\bibitem{moor2023med}
Michael Moor, Qian Huang, Shirley Wu, Michihiro Yasunaga, Yash Dalmia, Jure Leskovec, Cyril Zakka, Eduardo~Pontes Reis, and Pranav Rajpurkar.
\newblock Med-flamingo: a multimodal medical few-shot learner.
\newblock In {\em Machine Learning for Health (ML4H)}, pages 353--367. PMLR, 2023.

\bibitem{ji2023survey}
Ziwei Ji, Nayeon Lee, Rita Frieske, Tiezheng Yu, Dan Su, Yan Xu, Etsuko Ishii, Ye~Jin Bang, Andrea Madotto, and Pascale Fung.
\newblock Survey of hallucination in natural language generation.
\newblock {\em ACM Computing Surveys}, 55(12):1--38, 2023.

\bibitem{zhang2023siren}
Yue Zhang, Yafu Li, Leyang Cui, Deng Cai, Lemao Liu, Tingchen Fu, Xinting Huang, Enbo Zhao, Yu~Zhang, Yulong Chen, et~al.
\newblock Siren's song in the ai ocean: A survey on hallucination in large language models.
\newblock {\em arXiv preprint arXiv:2309.01219}, 2023.

\bibitem{lee2018hallucinations}
Katherine Lee, Orhan Firat, Ashish Agarwal, Clara Fannjiang, and David Sussillo.
\newblock Hallucinations in neural machine translation.
\newblock 2018.

\bibitem{maynez-etal-2020-faithfulness}
Joshua Maynez, Shashi Narayan, Bernd Bohnet, and Ryan McDonald.
\newblock On faithfulness and factuality in abstractive summarization.
\newblock In Dan Jurafsky, Joyce Chai, Natalie Schluter, and Joel Tetreault, editors, {\em Proceedings of the 58th Annual Meeting of the Association for Computational Linguistics}, pages 1906--1919, Online, July 2020. Association for Computational Linguistics.

\bibitem{pu2023summarization}
Xiao Pu, Mingqi Gao, and Xiaojun Wan.
\newblock Summarization is (almost) dead.
\newblock {\em arXiv preprint arXiv:2309.09558}, 2023.

\bibitem{zhou2020detecting}
Chunting Zhou, Graham Neubig, Jiatao Gu, Mona Diab, Paco Guzman, Luke Zettlemoyer, and Marjan Ghazvininejad.
\newblock Detecting hallucinated content in conditional neural sequence generation.
\newblock {\em arXiv preprint arXiv:2011.02593}, 2020.

\bibitem{rohrbach2018object}
Anna Rohrbach, Lisa~Anne Hendricks, Kaylee Burns, Trevor Darrell, and Kate Saenko.
\newblock Object hallucination in image captioning.
\newblock {\em arXiv preprint arXiv:1809.02156}, 2018.

\bibitem{shi2023replug}
Weijia Shi, Sewon Min, Michihiro Yasunaga, Minjoon Seo, Rich James, Mike Lewis, Luke Zettlemoyer, and Wen-tau Yih.
\newblock Replug: Retrieval-augmented black-box language models.
\newblock {\em arXiv preprint arXiv:2301.12652}, 2023.

\bibitem{lin2021truthfulqa}
Stephanie Lin, Jacob Hilton, and Owain Evans.
\newblock Truthfulqa: Measuring how models mimic human falsehoods.
\newblock {\em arXiv preprint arXiv:2109.07958}, 2021.

\bibitem{li2023evaluating}
Yifan Li, Yifan Du, Kun Zhou, Jinpeng Wang, Wayne~Xin Zhao, and Ji-Rong Wen.
\newblock Evaluating object hallucination in large vision-language models.
\newblock {\em arXiv preprint arXiv:2305.10355}, 2023.

\bibitem{tu2024towards}
Tao Tu, Shekoofeh Azizi, Danny Driess, Mike Schaekermann, Mohamed Amin, Pi-Chuan Chang, Andrew Carroll, Charles Lau, Ryutaro Tanno, Ira Ktena, et~al.
\newblock Towards generalist biomedical ai.
\newblock {\em NEJM AI}, 1(3):AIoa2300138, 2024.

\bibitem{hatem2023call}
Rami Hatem, Brianna Simmons, and Joseph~E Thornton.
\newblock A call to address ai “hallucinations” and how healthcare professionals can mitigate their risks.
\newblock {\em Cureus}, 15(9), 2023.

\bibitem{stanley2013logic}
Donald~E Stanley and Daniel~G Campos.
\newblock The logic of medical diagnosis.
\newblock {\em Perspectives in Biology and Medicine}, 56(2):300--315, 2013.

\bibitem{johri2023guidelines}
Shreya Johri, Jaehwan Jeong, Benjamin~A Tran, Daniel~I Schlessinger, Shannon Wongvibulsin, Zhuo~Ran Cai, Roxana Daneshjou, and Pranav Rajpurkar.
\newblock Guidelines for rigorous evaluation of clinical llms for conversational reasoning.
\newblock {\em medRxiv}, pages 2023--09, 2023.

\bibitem{gao2023llama}
Peng Gao, Jiaming Han, Renrui Zhang, Ziyi Lin, Shijie Geng, Aojun Zhou, Wei Zhang, Pan Lu, Conghui He, Xiangyu Yue, et~al.
\newblock Llama-adapter v2: Parameter-efficient visual instruction model.
\newblock {\em arXiv preprint arXiv:2304.15010}, 2023.

\bibitem{cai2023making}
Mu~Cai, Haotian Liu, Siva~Karthik Mustikovela, Gregory~P Meyer, Yuning Chai, Dennis Park, and Yong~Jae Lee.
\newblock Making large multimodal models understand arbitrary visual prompts.
\newblock {\em arXiv preprint arXiv:2312.00784}, 2023.

\end{thebibliography}

\newpage
\appendix
\renewcommand\thefigure{\thesection.\arabic{figure}} 
\renewcommand\thetable{\thesection.\arabic{table}} 
\counterwithin{figure}{section}

\appendix
\section{Data}

\subsection{Additional Medical Context}
In this paper, we focus on the analysis of bone marrow pathology slides for the diagnosis of blood cancer disorders. Specifically, our dataset contains 512x512 pixel images of 16,340 pathology patches corresponding to healthy, inconclusive, acute myeloid leukemia, and multiple myeloma cases.
The process for the analysis of bone marrow pathology slides, involves multiple steps. A pathologists has to first identify image regions that are deemed adequate for evaluation, excluding cases with either too low image quality or where the presence of other cells (e.g. red blood cells) prevents accurate medical diagnosis. Subsequently, the remaining adequate regions are examined to determine if they exhibit characteristics indicative of cancerous tissue. Specifically, in bone marrow aspirates, the assessment focuses on whether there is abnormal proliferation of cells in the regions of interest. Depending on the type of cells proliferating, the patient may be diagnosed with a corresponding hematological disorder. For instance, in our dataset, the uncontrolled proliferation of blast cells is indicative of acute myeloid leukemia, while similar proliferation of plasma cells suggests multiple myeloma.

\subsection{Generating a multi-turn conversation dataset}
The below section provides further details on the steps we took to derive a multimodal multi-turn conversation dataset in this specialist problem domain.

\textbf{Image data}: We obtained whole pathology slide images sourced from the clinical archives of an academic medical center. These were then segmented into 512x512 pixel patches\footnote{During training, we resize the image to a resolution of 256x256 pixels before feeding it into the image encoder.} and labelled as either "adequate for analysis", "particle-enriched contamination" or "blood contamination" after pathologist review. Subsequently we leveraged specialist software in order to obtain cell-counts, based on which a pathologist labelled cases with a high increase in blast or plasma cells as acute myeloid leukemia or multiple myeloma, respectively. Table~\ref{tab:diagnosis} details the distribution of the final diagnosis corresponding to the image data.

\textbf{Question Answer generation:}
Next, we utilize the symbolic representation of the bone marrow pathology slide analysis process to create clinically meaningful multi-turn conversations. This is accomplished by filling in question and answer templates based on the respective label for each analysis step. To prevent our model from overfitting to specific expressions in these templates, we increased the diversity of questions and answers by obtaining multiple question templates from our clinical collaborators and using GPT-4 to paraphrase these templates. The respective prompts are provided below:

$\textbf{Prompt for question paraphrasing:}$
"Perform $X$ times augmentation of the following sentence, it is for medical questions so make sure you preserve the meaning concisely."

$\textbf{Prompt for answer paraphrasing}:$ 
"Perform $X$ times augmentation of the following sentence, it is for medical diagnosis so make sure you preserve the meaning concisely: '{sentence}'. Also note that the question is '{question}', also don't repeat anything related to in response to the question, just make sure the single sentence is grammatically correct and makes sense."

\begin{table}[h!]
    \centering
    \caption{Distribution of Final Diagnoses in the Pathology Slide Image Dataset}
    \begin{tabular}{|l|r|}
        \hline
        Diagnosis & Number \\
        \hline
        Blood contamination  & 10083 \\
        Particle enriched contamination & 3510 \\
        Acute myeloid leukemia & 1531 \\
        Multiple myeloma  & 932 \\
        Healthy & 284 \\
        \hline
    \end{tabular}
    \vspace{0.035in}
    \label{tab:diagnosis}
\end{table}

\section{Instruction tuning details}

\subsection{Multimodal Supervised Finetuning} 
Using the instruction tuning dataset $\mathcal{D} = \{(\mathcal{I}_i, \{X^t_i, Y^t_i\}_t)\}_i$ ,== a straightforward approach to adapt VLMs for the diagnostic task at hand is by applying supervised finetuning. To construct this baseline, we use the LLaVA architecture \cite{liu2023improved, sun2023aligning} and jointly instruction-tune a vision encoder and a pre-trained LLM using token-level supervision to derive a supervised fine-tuned (SFT) model $\pi_{\text{SFT}}^{\phi}$.  Following prior work \cite{liu2024visual, liu2023improved}, the model is trained based on the LLMs original autoregressive training objective, where, for an answer sequence of length $T$, we compute the probability of the target answer as 
\begin{equation}
p(Y^t_i | X^t_i, {I}_i) = \Pi^T_{t=1} \pi_{\text{SFT}}^{\phi} (y_j | \mathcal{I}_i, \{X^{t^\prime}_{i}, Y^{t^\prime}_{i}\}_{t^\prime<t})
\end{equation}
where $y_j$ refers to the current prediction token in the answer sequence and $\{X^{t^\prime}_{i}, Y^{t^\prime}_{i}\}_{t^\prime<t}$ refers to the tokens in the previous parts of the answer sequence. 

\subsection{VLM response labelling}

In this work we leverage a simple rule-based reward model that evaluates the correctness of LLM responses based on the presence of relevant keywords in their answer. The respective keywords are depicted in Table~\ref{app:keywords_reward_model}. For a certain keyword to be valid we require it to appear without negation. An answer is classified as 'no match' in case it does not contain any of the considered keywords for the respective analysis step. 

\begin{table}[h]
\centering
\renewcommand{\arraystretch}{1.5} 
\caption{Keywords considered in rule-based reward model}
\begin{tabular}{|>{\raggedright\arraybackslash}p{0.2\linewidth}|>{\raggedright\arraybackslash}p{0.3\linewidth}|>{\raggedright\arraybackslash}p{0.4\linewidth}|}
    \hline
    \textbf{Analysis Steps} & \textbf{Classification} & \textbf{Keywords} \\ \hline
    \multirow{2}{0.9\linewidth}{Image Quality Assessment} & High quality & effective, appropriate, suitable, sufficient, optimal \\
    & Low quality & not, no, inadequate, unsuitable \\ 
    & No Match & - \\ \hline
    \multirow{4}{0.9\linewidth}{Cell Quality Assessment} & Adequate & optimal, advantageous, suitable, adequate, well, prime \\
    & Blood & blood, RBC \\
    & Clot & particles \\
    & No Match & - \\ \hline
    \multirow{4}{0.9\linewidth}{Cell Abnormality Analysis} & Normal & normal, healthy, no abnormality \\
    & Abnormal & cancer, disorder, malignancy \\
    & Inadequate & low, subpar, inadequate \\
    & No Match & - \\ \hline
    \multirow{5}{0.9\linewidth}{Detailed Cell Proliferation Reasoning} & Blast Cell Proliferation & myeloblast \\
    & Plasma Cell Proliferation & plasma cells \\
    & Normal & no abnormal, no proliferation, normal \\
    & Inadequate & low, subpar, inadequate \\
    & No Match & - \\ \hline
    \multirow{5}{0.9\linewidth}{Final Diagnosis} & Healthy & no malignancy phenotype, healthy \\
    & Acute Myeloid Leukemia & acute myeloid leukemia, AML \\
    & Multiple Myeloma & multiple myeloma, MM \\
    & Inconclusive & low quality, inadequate \\
    & No Match & - \\ \hline
\end{tabular}
\label{app:keywords_reward_model}
\vspace{-.2in}
\end{table}

\subsection{Training details} 
As our study concentrates on the performance of the finetuning algorithm, we base Dr-LLaVA on the same model architecture as LLaVA \cite{liu2024visual}. Our LLM utilizes Vicuna-V1.5-7b \cite{touvron2023llama, touvron2023llama-b, chiang2023vicuna}, paired with the pre-trained CLIP visual encoder ViT-L/14 at an image resolution of $256 \times 256$ \cite{radford2021learning}. Grid features are employed both before and after the final transformer layer to enhance the model's integration of visual data. We use a linear layer to map image features into the word embedding space, drawing on the pre-trained linear projection matrix checkpoints from LLaVA. We then conducted supervised fine-tuning for four epochs.

During the RL phase, following \cite{dubois2023alpacafarm} and \cite{sun2023aligning}, we initialized the value model based on the LLavA-13B-based reward model. We used LoRA-based finetuning with a rank of 64 for both the attention and feed-forward network modules. Consistent with \cite{dubois2023alpacafarm}, we used a batch size of 512 and normalized the advantage across the batch for each PPO step. The peak learning rate was set at $3 \times 10^{-5}$, applying cosine decay, and gradients were clipped by their Euclidean norm with a threshold of 1. Training was conducted through four complete rounds using our held-out RL data. For generalized advantage estimation, we set both $\lambda$ and $\gamma$ to 1, and adopted a constant KL regularizer coefficient of 0.1. The \textbf{\texttt{Dr-LLaVA}} model was trained using four A100 80 GB GPUs.

We leverage 80\% of our synthesized clinical multi-turn conversation dataset for supervised finetuning and RL and use the remaining 20\% for evaluation. We split the data at the conversation level such that all question-answer pairs pertaining to a particular image belong to the same sample. We use different prompt templates and rephrasing for the question-answer pairs in the training and testing sets to ensure that the models do not over-fit to specific phrasing of the clinician-VLM conversations

\clearpage
\section{Additional Experimental Results}

\subsection{Evaluation metrics}
To effectively assess the performance of our proposed model, we measure the accuracy of our model at the question, conversation and diagnosis level.

\begin{enumerate}
    \item \textbf{Question-level Accuracy ($A_Q$):} This metric evaluates the model's performance at the single question level. It is obtained by dividing the number of questions answered correctly by the total number of questions:
    \begin{equation}
        A_Q = \frac{\text{Number of correct answers}}{\text{Total number of questions}}
    \end{equation}

    \item \textbf{Conversation-level Accuracy ($A_C$):} This metric assesses the model's accuracy at the conversation level. Here we only consider a VLM's response as correct if it is able to correctly answer all questions pertaining to a multi-turn conversation about a specific case.
    \begin{equation}
        A_C = \frac{\text{Number of conversations with all questions answered correctly}}{\text{Total number of cases}}
    \end{equation}
    This metric asses the model's capability to consistently provide accurate answers across all questions within a multi-turn conversation, enabling the model to be a trustworthy companion throughout the full image analysis process.

    \item \textbf{Diagnostic Accuracy ($A_D$):} This metric focuses solely on the VLMs' responses to questions about the final diagnosis, as this is often the primary concern for medical decision-makers:
    \begin{equation}
        A_D = \frac{\text{Number of correctly answered diagnosis questions}}{\text{Total number of cases}}
    \end{equation}
\end{enumerate}

In conclusion, these three distinct levels of accuracy—$A_Q$, $A_C$, and $A_D$—provide a comprehensive evaluation of the proposed model's effectiveness in handling different aspects of medical inquiries. By breaking down the analysis to question, conversation, and diagnosis levels, we can better understand the model's strengths and pinpoint areas for improvement in handling complex medical scenarios.

\newpage

\subsection{Performance given diverse conversation sequences}

To capture the diverse forms of possible interactions between clinicians and VLMs, we assessed all VLMs using 3 conversational scenarios: \textbf{(1) Standard~Interaction~(SI)}: adheres to the logical dialogue sequence in Fig. \ref{fig:logic}, starting with image quality assessment and advancing through morphological analysis to reach a final diagnosis; \textbf{(2) Diagnosis First (DF)}: inverts the sequence in Fig. \ref{fig:logic}, where the clinician starts by asking about the patient diagnosis and then interacts with the model to understand the reasoning behind it; \textbf{(3) Improvised Interaction~(II)}: mimics the unpredictability of real-world interactions by randomizing the question sequence, presenting questions in a non-linear and potentially repetitive sequence. This is implemented~by~randomly~sampling questions pertaining to specific conversation with replacement. 

Table~\ref{app-results-order} presents the comparative results. \textbf{\texttt{Dr-LLaVA}} significantly outperforms the baseline models in non-traditional sequences, with performance gains ranging from 4.1 to 12.5 percentage points. This superior performance underlines \textbf{\texttt{Dr-LLaVA}}'s advanced adaptive reasoning capabilities, allowing it to extract critical information from conversational contexts effectively, regardless of the question sequencing. The ability of our model to handle these varied conversational dynamics demonstrates its potential in realistic clinical settings where dialogues may not follow a predefined order.

\begin{table}[H]
  \caption{\textbf{\texttt{Dr-LLaVA}} Performance in multi-turn conversations with varying order}
  \label{app-results-order}
  \centering
    \begin{tabular}{@{}l@{\hspace{3pt}}l@{\hspace{3pt}}P{1cm}@{\hspace{3pt}}P{1cm}@{\hspace{3pt}}P{1cm}@{}}
    \toprule
    \multirow{2}{*}{{\bf Model}} & \multirow{2}{*}{{\bf Metric}} & \multicolumn{3}{c}{{\bf Experiments}} \\
    & & SI & DF & II \\
    \midrule 
    \multirow{3}{*}{MiniGPT-4-SFT} 
    & $A_Q$ &75.8 &66.2 &72.2   \\
    & $A_C$ &44.2& 40.8& 41.6   \\
    & $A_D$ & 75.4& 50.0 & 71.4 \\
    \midrule
    \multirow{3}{*}{LLaMA-Adapter-SFT} 
    & $A_Q$ & 70.4& 65.2& 66.4  \\
    & $A_C$ & 42.5& 40.2& 43.5  \\
    & $A_D$ & 75.4& 74.6& 70.0  \\
    \midrule
    \multirow{3}{*}{OpenFlamingo-SFT} 
    & $A_Q$ &81.4 &65.2 & 70.0  \\
    & $A_C$ & 46.4& 40.3& 41.2  \\
    & $A_D$ & 69.9& 55.2& 72.0  \\
    \midrule
    \multirow{3}{*}{LLaVA-Med-SFT} 
    & $A_Q$ &91.2 &86.2 &85.4   \\
    & $A_C$ &85.6& 70.8& 71.6   \\
    & $A_D$ & 90.3& 82.2 & 81.3 \\
    \midrule
    \multirow{3}{*}{LLaVA-SFT} 
    & $A_Q$ & 92.4 & 83.1 & 82.0  \\
    & $A_C$ & 90.1 & 67.5 & 74.6  \\
    & $A_D$ & 91.8 & 76.9 & 76.9  \\
    \midrule
    \multirow{3}{*}{\textbf{\texttt{Dr-LLaVA}}} 
    & $A_Q$ & \textbf{93.6} & \textbf{88.9} & \textbf{92.0}  \\
    & $A_C$ & \textbf{90.8} & \textbf{84.4} & \textbf{87.4}  \\
    & $A_D$ & \textbf{92.0} & \textbf{85.9} & \textbf{89.0}  \\
    \bottomrule
    \end{tabular}
\end{table}

\newpage

\subsection{Performance in case of misleading clinician hypothesis}

\subsubsection{Results}

We also evaluate the model's performance in scenarios where physicians incorporate hypotheses into their prompts. Specifically, we examine two types of queries: Confirmation Queries (CQ), where the clinician seeks model validation of their (potentially wrong) hypothesis, and Rationalization Queries (RQ), where~the~clinician presents a (potentially wrong) explanation for their hypothesis and asks the model about next diagnostic steps. The precise formulations for each query type are detailed in section \ref{app: misleading}.

Table~\ref{app-results-misleading} shows the accuracy of various VLMs in distinguishing between accurate and~misleading~information. (``R'' corresponds to clinician prompts with right information and ``W''~refers~to~wrong ones.) The performance of all models was robust when physician prompts included accurate hypotheses, but accuracy notably declined for all models when the prompts contained misleading information. In these scenarios, \textbf{\texttt{Dr-LLaVA}} consistently exhibited a higher rate of disagreement with the misleading content, suggesting that our alignment algorithm more effectively anchors the symbolic reasoning process (Fig. \ref{fig:logic}) in the visual data, thus enabling the model to detect erroneous textual inputs.

\begin{table}[H]
  \caption{\textbf{\texttt{Dr-LLaVA}} Performance under misleading clinician prompts}
  \centering
  \label{app-results-misleading}
    \begin{tabular}{@{}l@{\hspace{3pt}}l@{\hspace{3pt}}P{1cm}@{\hspace{3pt}}P{1cm}@{\hspace{3pt}}P{1cm}@{\hspace{3pt}}P{1cm}@{}}
    \toprule
    \multirow{2}{*}{{\bf Model}} & \multirow{2}{*}{{\bf Metric}} & \multicolumn{4}{c}{{\bf Experiments}} \\
    \cmidrule{3-6}
    & & CQ-R & CQ-W & RQ-R & RQ-W \\
    \midrule 
    \multirow{3}{*}{LLaVA-SFT} 
    & AQ & 99.2 & 13.8 & 91.5 & 33.3 \\
    & AD & 99.3 & 13.6 & \textbf{99.8} & 31.3 \\
    \midrule
    \multirow{3}{*}{\textbf{\texttt{Dr-LLaVA}}} 
    & AQ & 99.0 & \textbf{22.7} & \textbf{93.0} & 39.0 \\
    & AD & 97.9 & \textbf{33.7} & 98.6 & \textbf{48.6} \\
    \bottomrule
    \end{tabular}
\end{table}

\subsubsection{Example Prompts for Evaluating Model Performance with Clinician Hypotheses}
\label{app: misleading}

This section presents example prompts crafted to evaluate our model’s ability to respond to scenarios where physicians incorporate hypotheses into their prompts. The prompts are divided into two categories: Confirmation Queries (CQ) and Rationalization Queries (RQ).

\textbf{Confirmation Query (CQ) Prompts}

Confirmation Queries aim to assess the model's ability to validate clinician opinions. These queries challenge the model to either concur with or contest a clinician's judgment, which may be accurate (CQ-R) or erroneous (CQ-W).

\textbf{Example Prompt 1 (CQ-R):}
"After reviewing the image, the clinician believes that [correct statement]. Do you agree with this assessment?"

\textbf{Example Prompt 2 (CQ-W):}
"After examining the image, the clinician suggests that [misleading statement]. Do you concur with this opinion?"

\textbf{Rationalization Query (RQ) Prompts}

Rationalization Queries present the model with a previous conclusion, which may be correct (RQ-R) or incorrect (RQ-W), and ask about the next diagnostic steps. These queries assess the model's ability to correct incorrect hypotheses even when not explicitly prompted to do so.

\textbf{Example Prompt 3 (RQ-R):}
"A previous clinician reviewed the image and concluded that [accurate rationale]. Considering this, what would be your next step in the diagnostic process? [Question]"

\textbf{Example Prompt 4 (RQ-W):}
"A previous clinician interpreted the image and believed [erroneous rationale]. With this in mind, how would you proceed with the diagnosis? [Question]"
\newpage

\subsection{The correctness and consistency trade-off}

The hyperparameter $\lambda$ in (\ref{reward_eq})~balances~between the model correctness on responses to individual question and the overall alignment of these responses with a valid reasoning process across a conversation. Setting $\lambda$ to a large value heavy regularizes the conversational output and may~encourage~the model to follow valid reasoning processes that are not grounded in the input image, e.g., the model could always follow a $\mbox{\footnotesize (Low image quality $\to$ Inconclusive)}$ symbolic rule regardless of the input image. On the other hand, setting $\lambda=0$ reduces to the standard supervised finetuning setup where~the~model optimizes for question-level accuracy but is likely to exhibit context-conflicting hallucinations within conversations. 

We define context-conflicting hallucinations as an answer that deviates from the expected pathways outlined in the symbolic representation of the pathology slide analysis process, as illustrated in \ref{fig:logic}. We use a rule-based labelling model to classify the VLM responses according to the possible choices within the symbolic representation of medical reasoning. This allows us to quantify the proportion of answers that do not follow any logical trajectory within this framework.

\begin{figure}[H]
\centering
\includegraphics[width=2.in]{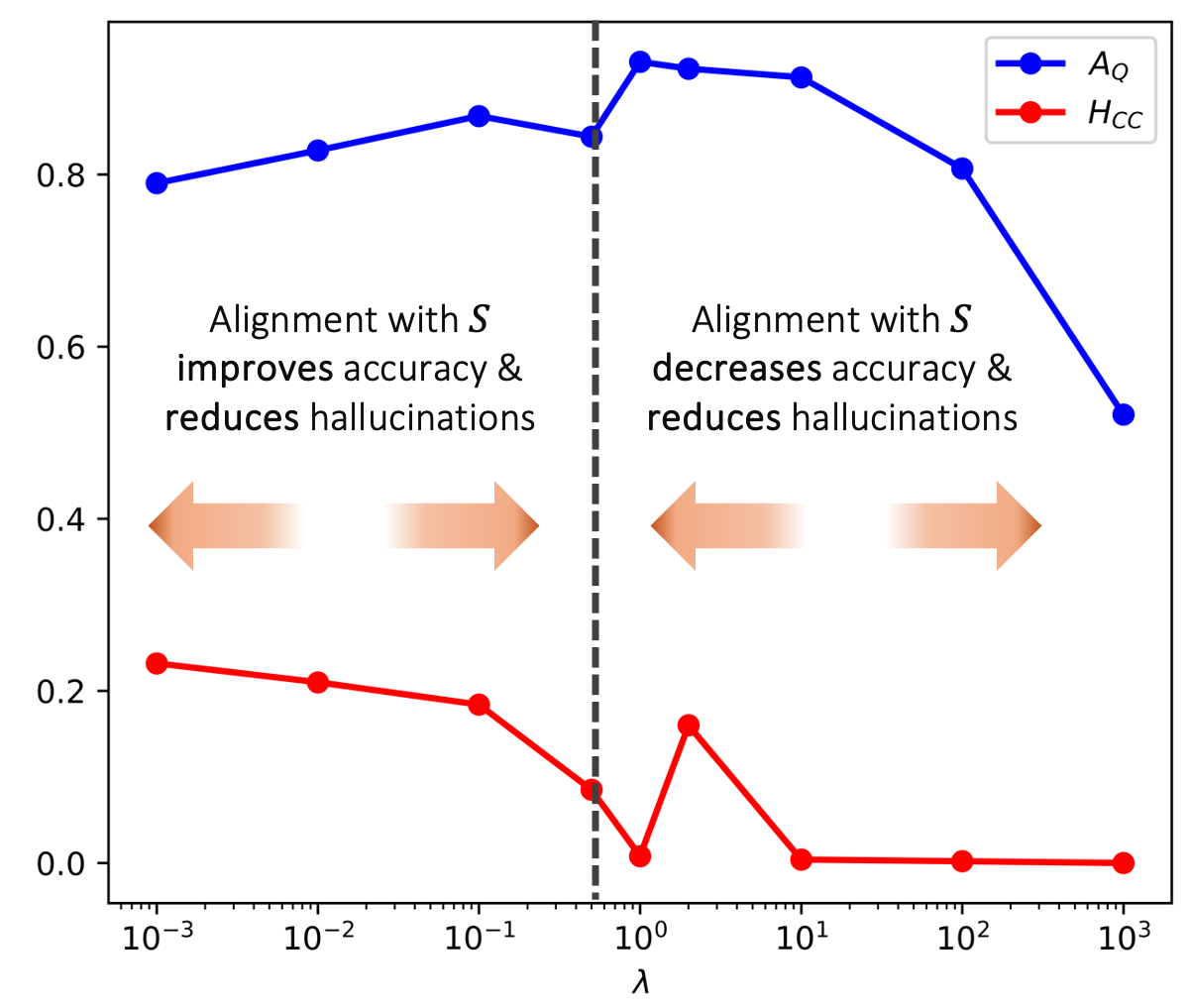}
\vspace{-.1in}
\caption{Impact of the hyperparameter $\lambda$ on Dr-LLaVA performance.}
\label{Fig:lambda}
\end{figure}

Fig.~\ref{Fig:lambda} demonstrates the impact of the choice of $\lambda$ on the model performance in terms of $A_Q$ and the corresponding rate of context-conflicting hallucinations $H_{cc}$. Here, we define $H_{cc}$ as the fraction of conversations that map to invalid symbolic rules, i.e., $H_{cc} = E[\boldsymbol{1}\{\hat{s} \notin \mathcal{S}\}]$. The plot shows that increasing $\lambda$ initially improves accuracy and consistency (quantified through $H_{cc}$) reaching an optimal point beyond which further increases in $\lambda$ lead to diminished accuracy. These findings show that our alignment with valid clinical reasoning not only improves the model's coherence and trustworthiness, but can also improve the model accuracy on individual questions by regularizing the entire conversational output using prior knowledge on diagnostic scenarios.

\newpage

\subsection{Ablation study of different reward model components}

\begin{figure}[H]
\centering
  \includegraphics[width=5.5in]{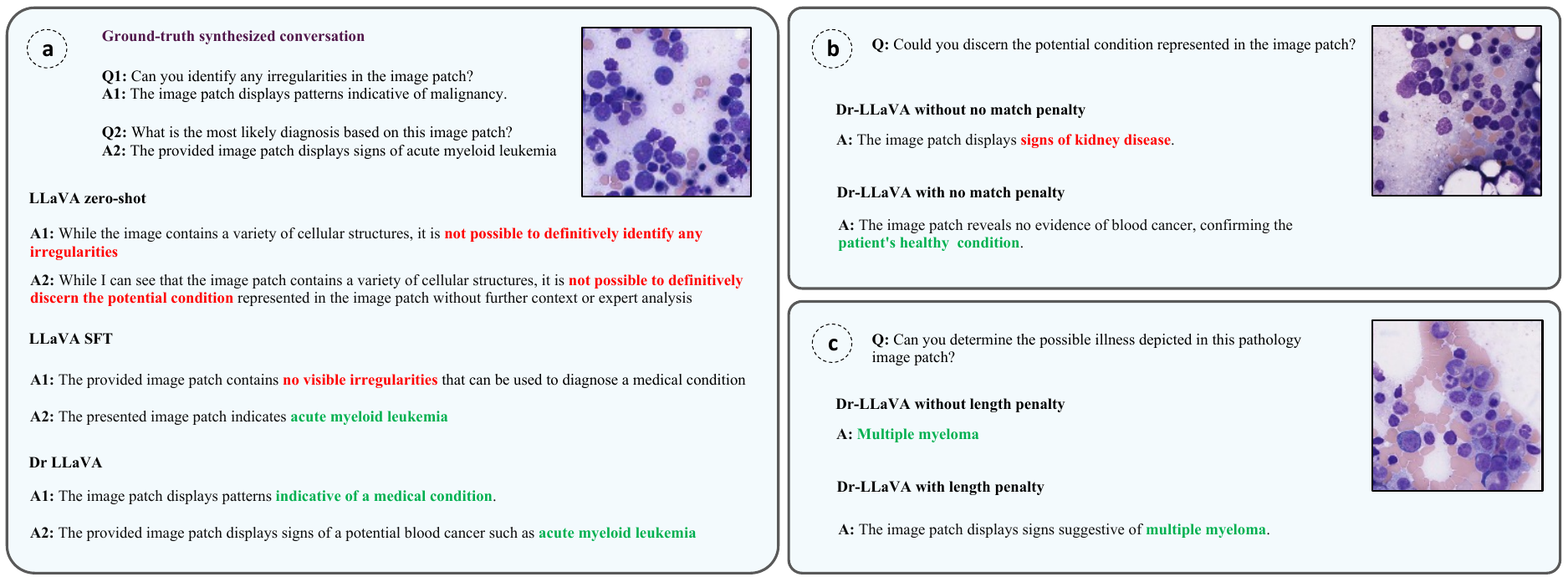}
  \vspace{-.15in}
  \caption{Example outputs of the \textbf{\texttt{Dr-LLaVA}} model and baselines.}
  \label{fig:Dr-LLaVA-framework2}
  \vspace{.025in}
  \vspace{-.25in}
\end{figure}

We examined the impact of excluding various components of the reward model in (\ref{reward_eq}). Our findings, shown in Table~\ref{app-rew-comps}, indicate that omitting either the correctness or consistency rewards significantly reduces predictive accuracy. As expected, removing the correctness reward ($R_c$) improves answer consistency. This occurs because the model is then primarily driven to align with abstract reasoning, disregarding the actual correctness of the responses in the context of the visual input. Eliminating the length penalty ($R_l$) and no-match penalty ($R_m$) resulted in moderate yet noticeable declines in both accuracy and consistency. Qualitatively, the absence of these penalties demonstrate their vital role in preventing reward hacking and maintaining the integrity of medical dialogue. For instance, the removal of~the~no-match~penalty caused a marked deterioration in content relevance and accuracy, with the model occasionally generating blatantly unrelated medical suggestions. An example of this is the~inappropriate~reference~to renal conditions when analyzing bone marrow images (Fig. \ref{fig:Dr-LLaVA-framework2}(b)). Additionally, without the~length~penalty, the model tended towards producing brief, often incomplete responses as observed in Fig. \ref{fig:Dr-LLaVA-framework2}(c).

\begin{table}[H]
  \caption{Ablation study of \textbf{\texttt{Dr-LLaVA}} reward model components}
  \label{app-rew-comps}
  \centering
    \begin{tabular}{@{}lcccc@{}}
        \toprule
        {\bf Scenarios} & \multicolumn{2}{c}{{\bf Single-turn VQA}} & \multicolumn{2}{c}{{\bf Multi-turn VQA}} \\
          \cmidrule(lr){2-3} \cmidrule(lr){4-5}
         & $A_Q$ & $H_{cc}$ & $A_Q$ & $H_{cc}$ \\
        \midrule
         \textbf{\texttt{Dr-LLaVA}} &\textbf{89.6} &22.5 &\textbf{92.5} &5.4 \\
         \textbf{\texttt{Dr-LLaVA}} w/o $R_c$ &32.4& \textbf{1.5} &52.1 &\textbf{0.0} \\
         \textbf{\texttt{Dr-LLaVA}} w/o $R_S$ &78.4 &47.5 & 83.0 &20.2\\
         \textbf{\texttt{Dr-LLaVA}} w/o $R_m$ &85.2 &30.6 &87.6 &8.8 \\
         \textbf{\texttt{Dr-LLaVA}} w/o $R_l$ &87.9 &25.8 &89.1 &7.0 \\
         \textbf{\texttt{Dr-LLaVA}} w/o $R_m$/$R_l$ &84.2&33.1 &87.2 &10.1  \\
        \bottomrule
    \end{tabular}
\end{table}

\newpage

\section{Related Work}
{\bf Vision-Language Models for medicine.} Large Language Models (LLMs) \cite{brown2020language, openai_gpt4_2023, chowdhery2023palm, anil2023palm, workshop2022bloom, muennighoff2022crosslingual, touvron2023llama, touvron2023llama-b, taori2023stanford, chiang2023vicuna} have excelled in generating high-quality textual responses across diverse tasks, fueling advancements in chat-based AI assistants \cite{openai_chatgpt, unknown-author-2023}. Recent work has extended these models to handle multimodal image-text data \cite{su2019vl, ramesh2021zero, alayrac2022flamingo}, which has led to the emergence of powerful vision-language models (VLM) including OpenFlamingo \cite{awadalla2023openflamingo}, MiniGPT-4 \cite{zhu2023minigpt} and LLaVA \cite{li2023llava}. In the medical domain, the integration of images and texts has been explored in areas such as ultrasound \cite{byra2023few, liu2023qilin}, pathology \cite{huang2023visual, lu2023towards}, and radiology \cite{huang2021gloria, wu2023towards}, typically utilizing modality-specific vision encoders. Additionally, recent studies have proposed models that directly finetune state-of-the-art VLMs for medical applications including Med-Alpaca \cite{shu2023visual}, Med-Flamingo \cite{moor2023med} and LLaVAMed \cite{li2023llava}.~However,~these~models solely leverage instruction-tuning with token-level supervision, but do~not~consider~regularizing the model outputs on a conversation-level by incorporating domain knowledge on diagnostic pathways.

{\bf Hallucination in generative models.} In the Natural Language Processing (NLP) literature, ``hallucination'' was defined as the phenomenon where a model generates content diverging from the original source material \cite{ji2023survey}. With the advent of advanced LLMs, this definition has expanded. As noted in \cite{zhang2023siren}, hallucination can manifest in three distinct ways: 1) {\it Input-conflicting} hallucination, observed in scenarios like machine translation and summarization, where the model's response alters or misinterprets the static context of the user's prompt \cite{lee2018hallucinations, maynez-etal-2020-faithfulness, pu2023summarization, zhou2020detecting}; 2) {\it Context-conflicting} hallucination, where the model's output contradicts its previous responses \cite{rohrbach2018object,shi2023replug}; and 3) {\it Fact-conflicting} hallucination, in which the generated content conflicts with established factual knowledge \cite{lin2021truthfulqa, li2023evaluating}.

To the best of our knowledge, our finetuning framework is the first to address context-conflicting hallucinations, which are particularly important in clinical applications \cite{shi2023replug, tu2024towards, hatem2023call}. This is because medical practitioners adhere to stringent logical processes in diagnosis and avoid conclusions that contradict previous observations \cite{stanley2013logic}. Therefore, a VLM that accurately identifies the final diagnosis but fails to correctly respond to preceding observation-related questions would~be~deemed~unreliable~\cite{johri2023guidelines}. Similar to prior work, our reward model in (\ref{reward_eq}) addresses input- and fact-conflicting~hallucination,~but is distinguished by inclusion of the symbolic reward $R_{\mathcal{S}}$ to address context-conflicting hallucination.


{\bf Addressing misalignment in Vision-Language Models.} The two predominant methods for aligning VLM outputs with specific domain requirements or general human preferences are supervised finetuning and Reinforcement Learning from Human Feedback (RLHF). Similar to LLMs, supervised finetuning typically involves training a pre-trained model on a dataset tailored to the task at hand \cite{gao2023llama, zhang2023llama, liu2023improved, zhang2023llavar, cai2023making}. However, this approach can lead to misalignment between image and text modalities in VLMs, resulting in outputs insufficiently grounded in the visual context \cite{sun2023aligning}. Conversely, RLHF has proven effective in recalibrating models to match human preferences. This method relies on preference data from human labelers to train a reward model, which then finetunes the VLM using reinforcement learning techniques such as Proximal Policy Optimization (PPO) \cite{schulman2017proximal}. In our work, which, to the best of our knowledge, is the first to apply RL-based finetuning to VLMs for the medical domain, we introduce a new RL framework tailored to the medical decision-making contexts by using an automatic reward function to reduce the reliance on expensive specialist annotators.


\end{document}